\pdfoutput=1
\documentclass{article}
\usepackage{spconf,amsmath,amssymb,graphicx}
\usepackage{caption}
\usepackage{subcaption}
\usepackage{fancyhdr}
\usepackage{float}

\def\x{\mathbf{x}}
\def\y{\mathbf{y}}
\def\p{\mathbf{p}}
\def\z{\mathbf{z}}
\def\k{\mathbf{k}}
\def\h{\mathbf{h}}
\def\c{\mathbf{c}}

\def\yhat{\mathbf{\hat{y}}}
\def\henc{\mathbf{h}^\text{enc}}
\def\blank{\left<b\right>}
\def\sos{\left<\text{sos}\right>}
\def\na{\left<\text{n/a}\right>}
\def\eokw{\left<\text{eokw}\right>}
\def\eow{\left<\text{eow}\right>}
\def\spc{\left<\text{space}\right>}

\newcommand{\norm}[1]{\lvert #1 \rvert}

\fancypagestyle{FirstPage}{
  \fancyhf{}  
  \fancyfoot[L]{To appear in \textit{Proceedings of IEEE ASRU 2017}.}
}

\title{STREAMING SMALL-FOOTPRINT KEYWORD SPOTTING USING SEQUENCE-TO-SEQUENCE MODELS}
\name{Yanzhang He, Rohit Prabhavalkar, Kanishka Rao, Wei Li, Anton Bakhtin, Ian McGraw}
\address{Google Inc.,\\
	Mountain View, CA, U.S.A.\\
	\texttt{\{yanzhanghe,prabhavalkar,kanishkarao,mweili,bakhtin,imcgraw\}@google.com}}
\begin{document}
\ninept
\maketitle
\thispagestyle{FirstPage}  

\begin{abstract}
	We develop streaming keyword spotting systems using a recurrent neural
	network transducer (RNN-T) model: an all-neural, end-to-end
	trained, sequence-to-sequence model which jointly learns acoustic and
	language model components.
	Our models are trained to predict either phonemes or graphemes as
	subword units, thus allowing us to detect arbitrary keyword phrases,
	without any out-of-vocabulary words.
	In order to adapt the models to the requirements of keyword spotting,
	we propose a novel technique which biases the RNN-T system towards a
	specific keyword of interest.

	Our systems are compared against a strong sequence-trained,
	connectionist temporal classification (CTC) based ``keyword-filler''
	baseline, which is augmented with a separate phoneme language model.
	Overall, our RNN-T system with the proposed biasing technique
	significantly improves performance over the baseline system.
\end{abstract}

\begin{keywords}
	Keyword spotting, sequence-to-sequence models, recurrent neural network
transducer, attention, embedded speech recognition.
\end{keywords}

\section{Introduction}
\label{sec:introduction}
Keyword spotting (KWS), sometimes also referred to as spoken term detection, is
the task of detecting specific words, or multi-word phrases in speech
utterances.
Many previous works consider the problem of developing ``offline" (i.e.,
non-streaming) KWS technologies.
In this setting, the dominant paradigm consists of recognizing the entire speech
corpus using a large vocabulary continuous speech recognizer (LVCSR) to build
word or sub-word lattices, which can then be indexed to perform efficient
search, e.g.,~\cite{FiscusAjotGarofolo07, MillerKleberKaoEtAl07,
VergyriShafranStolckeEtAl07}.

In contrast to the methods described above, there is growing interest in
building ``online" (i.e., streaming) KWS systems which can be deployed on
mobile devices which are significantly limited in terms of memory and
computational capabilities.
In such applications, \emph{when deployed for inference}, the KWS system must
continuously process incoming audio, and only trigger when a specific keyword is
uttered.
In order to simplify the problem further, most previous works assume that the
model will only be required to detect a small number of possible keywords, thus
allowing the development of keyword-specific models.
Many previous works propose to train neural networks to identify word targets in
individual keywords: for example, using feed-forward deep neural
networks~\cite{ChenParadaHeigold14,
PrabhavalkarAlvarezParadaEtAl15, TuckerWuSunEtAl16}, convolutional
networks~\cite{SainathParada15} or recurrent neural
networks~\cite{SunRajuTuckerEtAl16, ArikKlieglChildEtAl17,
FernandezGravesSchmidhuberEtAl07}.
Such systems assume the availability of a large number of examples of the
keywords of interest in order to train models robustly.
Prominent examples of such technologies include speech-enabled assistants such
as ``Okay/Hey Google" on Google Home~\cite{LiSainathCaroselliEtAl17}, ``Alexa"
on the Amazon Echo, and ``Hey Siri" on Apple devices.
There has also been some prior work which has explored building low-footprint
KWS systems which can detect arbitrary keywords in the
incoming speech: for example, using structured support vector
machines~\cite{KeshetGrangierBengio09, PrabhavalkarLivescuFosler-LussierEtAl13},
and techniques based on matching incoming audio to example templates of the
keyword (Query-by-Example)~\cite{HazenShenWhite09, ChenParadaSainath15}.

Recently, end-to-end trained, sequence-to-sequence models have become popular
for speech recognition. Examples of such models include
the recurrent neural network transducer (RNN-T)~\cite{Graves12,
GravesMohamedHinton13}, the recurrent neural
aligner~\cite{SakShannonRaoEtAl17}, connectionist temporal
classification (CTC)~\cite{GravesFernandezGomezEtAl06} with
grapheme~\cite{HwangLeeSung15, LengerichHannun16},
syllable~\cite{BaiYiNiEtAl16} or word
targets~\cite{SoltauLiaoSak16}, and attention-based
models~\cite{ChanJaitlyLeEtAl16, BahdanauChorowskiSerdyukEtAl16,
LuZhangRenals16}.
Such models combine the acoustic, and language model components of a
traditional speech recognition system into a single, jointly trained model.
In recent work, we have shown that RNN-T and attention-based models, trained
on $\sim$12,500 hours of transcribed speech data to directly predict grapheme
sequences without a separate language model, perform competitively on dictation
test sets when compared against a state-of-the-art, discriminatively
sequence-trained, context-dependent phone-based recognizer, augmented with a
large language model~\cite{PrabhavalkarRaoSainathEtAl17}.
We have also shown, that sequence-to-sequence models trained to predict
phoneme-based targets, can be effective when used in a second pass rescoring
framework~\cite{PrabhavalkarSainathLiEtAl17}.

There has been some recent work which has explored sequence-to-sequence models
in the context of KWS. Zhuang et al.~\cite{ZhuangEtAl16} use a long short-term
memory (LSTM)~\cite{HochreiterSchmidhuber97} network with CTC to train a KWS
system that generates phoneme lattices for efficient search.
Rosenberg et al.~\cite{RosenbergAudhkhasiSethyEtAl17} apply attention-based
models to compute n-best lists of recognition results which are then indexed for
efficient search; performance, however, was found to be worse than a traditional
lattice-based KWS approach.
Audhkhasi et al.~\cite{AudhkhasiRosenbergSethyEtAl17} train an end-to-end system
to predict whether a given keyword (represented as a grapheme string) is present
in the speech utterance without explicitly decoding utterances into output
phoneme or word strings.

In the present work, we explore the use of sequence-to-sequence models,
specifically, RNN-T, to build a streaming KWS system which can be used to detect
arbitrary keywords.
Unlike a number of previous works which have only examined sequence-to-sequence
models in the context of graphemes, we train RNN-T systems to predict graphemes
as well as phonemes as sub-word units.
Additionally, we propose a novel technique to
\emph{bias the search towards a specific keyword of interest} using an
attention mechanism (described in more detail in Section~\ref{sec:attn}).
We find that RNN-T system trained to predict phonemes, when augmented with an
additional ``end-of-word" symbol (see Section~\ref{sec:keyword_filler})
strongly outperforms a
strong keyword-filler baseline derived from a sequence-trained CTC-based
recognizer~\cite{McGrawPrabhavalkarAlvarezEtAl16}.
Overall, our best performing system achieves a false reject (FR) rate of 8.9\%
at 0.05 false alarms (FA) per hour, compared to the baseline which achieves
14.5\% at the same FA threshold, which corresponds to a 39\% reduction in
the FR rate.

The organization of the rest of the paper is as follows. In
Section~\ref{sec:models} we describe various modeling strategies used in this
paper. Section~\ref{sec:baselines} describes our baseline approaches for
keyword spotting. We present our experimental setup in
Section~\ref{sec:experiments}, and discuss our results in
Section~\ref{sec:results}, before concluding in Section~\ref{sec:conclusions}.

\section{Modeling Strategies}
\label{sec:models}
In subsequent sections, we denote a sequence of parameterized acoustic features
as, $\x = \left[\x_1, \cdots, \x_T \right]$,
where, $\x_t \in \mathbb{R}^d$; $T$ denotes the number of acoustic frames in the
utterance.
We denote the corresponding sequence of output targets (e.g., graphemes or
phonemes) corresponding to the utterance as $\y = \left[ y_1, \cdots, y_L
\right]$, where, $y_i \in \mathcal{Y}$.
In the context of ASR, the input label sequence is typically much longer
than the target label sequence, i.e., $T > L$.

\begin{figure}
	\includegraphics[width=\columnwidth]{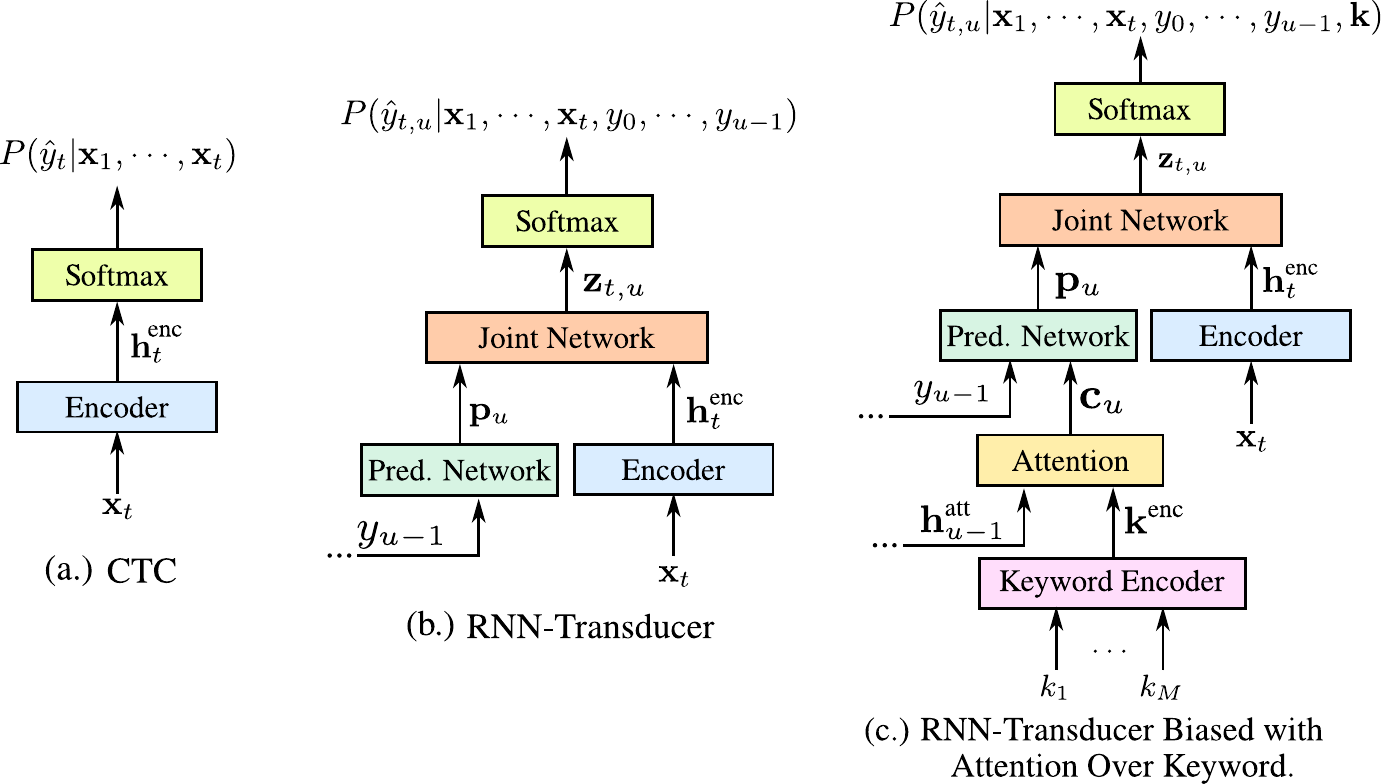}
	\caption{A schematic representation of the models used in this work.}
  	\label{fig:model-schematic}
	\vspace{-0.2in}
\end{figure}
\subsection{Connectionist Temporal Classification}
\label{sec:ctc}
CTC~\cite{GravesFernandezGomezEtAl06} is a technique for modeling a conditional
probability distribution over sequence data, $P(\mathbf{y} | \mathbf{x})$, when
frame-level alignments of the target label sequence are unknown.
CTC augments the set of output targets with an additional symbol, referred to as
the \emph{blank} symbol, denoted as $\blank$.
We denote by $\yhat = \left[\hat{y}_1, \cdots, \hat{y}_T\right] \in
\mathcal{B}(\x, \y)$, the set of all label sequences of length $\norm{\x} = T$,
such that
$\yhat_t \in \left\{ \mathcal{Y} \cup \blank \right\}$, for $1 \leq t \leq T$,
which are equivalent to $\y$ after first removing consecutive identical
symbols, and then removing any blank symbols: e.g.,
$xx{\blank}{\blank}y{\blank} \to xy$.

CTC models the output probability of the target sequence, $\y$, conditioned on
the input, $\x$, by marginalizing over all possible frame-level alignments,
where each output label is assumed to be independent of the other labels,
conditioned on $\x$:
\begin{equation}
	P(\y | \x) = \sum_{\yhat \in \mathcal{B}(\x, \y)} P(\yhat | \x) =
	\sum_{\yhat \in \mathcal{B}(\x, \y)} \prod_{t=1}^{T} P(\hat{y}_t | \x_1, \cdots, \x_t)
\end{equation}
The conditional probability, $P(\hat{y}_t | \x_1, \cdots, \x_t)$, can be computed using a
recurrent neural network (which we refer to as the \emph{encoder} network), as
illustrated in Figure~\ref{fig:model-schematic}(a.).
As shown in the figure, the encoder maps each input frame, $\x_t$, into a
higher-level representation, $\henc_t$, followed by a softmax layer which
converts $\henc_t$ into a probability distrubution $P(\hat{y}_t | \x_1, \cdots,
\x_t)$ over the output labels in $\left\{ \mathcal{Y} \cup \blank \right\}$.
The model can be trained using stochastic gradient descent to optimize
likelihood over the training set, given paired input and target sequences $(\x,
\y)$.
The gradients required for this process can be computed using the forward-backward
algorithm~\cite{GravesFernandezGomezEtAl06}.

\subsection{RNN Transducer}
\label{sec:rnnt}
Although CTC has been used successfully in many previous works in the context of
ASR (e.g.,~\cite{SakSeniorRaoEtAl15, MiaoGowayyedMetze15, SoltauLiaoSak16}), it
makes a strong conditional independence assumption since it assumes that outputs
at each step are independent of the history of previous predictions.
The RNN-T model improves the CTC approach by augmenting it with an additional
\emph{prediction network}~\cite{Graves12, GravesMohamedHinton13}, which is
explicitly conditioned on the history of previous outputs, as illustrated in
Figure~\ref{fig:model-schematic}(b.).
The RNN-T model may be viewed as a type of sequence-to-sequence model
architecture~\cite{ChanJaitlyLeEtAl16, BahdanauChorowskiSerdyukEtAl16}, where
the encoder (referred to as a \textit{transcription network} in~\cite{Graves12})
corresponds to the RNN acoustic model in a traditional recognizer, and the
\emph{prediction network} (together with the \emph{joint network}) corresponds
to the \emph{decoder}.
The decoder network may be viewed as an RNN language model which attempts to predict the
current label given the history of labels.
We note that unlike most attention-based models that have been explored in the
past (e.g.,~\cite{ChanJaitlyLeEtAl16, BahdanauChorowskiSerdyukEtAl16}), output
targets can be extracted from the RNN-T in a streaming fashion, since
the model does not have to examine the entire encoded utterance in order to
compute an output target label.

The prediction network is provided with the previous non-blank input label, $y_u
\in \mathcal{Y}$, as input, and produces a single output vector, denoted as
$\p_u$.
The prediction network is fed a special symbol at the start of decoding, $y_0 =
\sos$, which denotes the start of the sentence.

The joint network consists of a set of feed-forward layers which compute logits
$\z_{t,u}$ for every input frame $t$ and label $u$, using additional parameters
$A, B, b, D, d$, as follows:
\begin{align}
	\mathbf{h}^{\text{joint}}_{t, u} &= \tanh(A \mathbf{h}^\text{enc}_t + B \mathbf{p}_u + b) \\
	\mathbf{z}_{t, u} &= D \mathbf{h}^{\text{joint}}_{t, u} + d
\end{align}
These logits are passed to a final softmax layer which computes probabilities
over targets in $\left\{ \mathcal{Y} \cup \blank \right\}$.\footnote{These
equations correspond to Eq. 15--18 in~\cite{GravesMohamedHinton13}.}

The model can be trained to optimize likelihood over the training set, by
marginalizing over all possible alignments (i.e., $\mathcal{B}(\x, \y)$) similar
to CTC, using stochastic gradient descent where the required gradients are
computed using the dynamic programming algorithm described in~\cite{Graves12,
GravesMohamedHinton13}.

\subsection{Biasing the RNN-Transducer with the keyword of interest using the
attention mechanism}
\label{sec:attn}
Previous works that have examined the use of sequence-to-sequence models for KWS
(e.g., ~\cite{RosenbergAudhkhasiSethyEtAl17}) have typically only done so
indirectly; the models is trained for ASR, and used to generate n-best lists
which can be indexed for efficient search.
A notable exception, is work by Audhkhasi et
al.~\cite{AudhkhasiRosenbergSethyEtAl17} where the model is trained directly for
the KWS task which is similar to the query-by-example approach that has been
investigated previously~\cite{HazenShenWhite09}.

With the goal of improving KWS performance, we extend the RNN-T system described
in Section~\ref{sec:rnnt} with an attention-based keyword biasing mechanism in
the prediction network to make the model aware of the keyword of interest during
the search process.
This model can be thought of as a variant of the RNN-T model augmented with
attention, proposed in our previous work~\cite{PrabhavalkarRaoSainathEtAl17},
wherein we replace the prediction network with an attention-based decoder that
computes attention over the targets in the keyword phrase.
The intuition is that during inference, when the suffix of the current predicted
label sequence is close to the prefix of the keyword, the attention vector is
activated in the corresponding position within the keyword.
This, in turn, generates a context vector to bias the network prediction towards
the remaining part of the keyword.
Critically, since the keyword phrase only consists of a small number of targets,
the use of attention over the keyword does not introduce any latency or
significant computational overhead during inference.
This model is depicted in Figure~\ref{fig:model-schematic}(c.).

Specifically, at each step, the prediction network recieves, in addition to the
previous non-blank label $y_{u-1}$, a \emph{context vector}, $\c_u$ which is
computed using dot-product attention~\cite{ChanJaitlyLeEtAl16} over the keyword targets (phoneme targets, in our experiments).
We denote the sequence of phoneme targets in the keyword phrase to be detected,
as $\k = \left[k_1, \cdots, k_M, k_{M+1}\right]$, where
$M$ is the number of targets in the keyword phrase, and $k_{M+1}$ is a special
target that corresponds to ``not applicable", denoted $\na$.\footnote{We also
experimented with excluding this symbol, and only using the targets in the
keyword, and found that the overall performance was similar to a model with this
target. In this work, we only present results with the $\na$ keyword target.}
The keyword encoder takes as input the phoneme sequence, and outputs a matrix
$\mathbf{k}^\text{enc} = \left[k^\text{enc}_1, \cdots, k^\text{enc}_M,
k^\text{enc}_{M+1}\right]$, where $k^\text{enc}_i$
is a one-hot embedding vector of $k_i$, and $k^\text{enc}_{M+1}$ is a zero vector.
If we denote the state of the prediction network after predicting $u-1$ labels
as $\h^\text{att}_{u-1}$, the context vector, $\c_u$ is computed as follows:
\begin{align}
	\beta_{j, u} &= \left< \phi(k^\text{enc}_j) , \psi(\h^\text{att}_{u-1})\right>\quad
	\quad \text{ for each } 1 \leq j \leq M+1 \\
	\alpha_{j, u} &= \frac{e^{\beta_{j, u}}}{\sum_{j'=1}^{M+1} e^{\beta_{j',
	u}}} \label{eq:att_weights} \\
	\c_u &= \sum_{j=1}^{M+1} \alpha_{j, u} k^\text{enc}_j
\end{align}
\noindent where, $\phi(\cdot)$ and $\psi(\cdot)$ represent linear embeddings,
and $\left<\cdot, \cdot\right>$ represents the dot product between two vectors.
Thus, the prediction network produces an output $\p_u$ conditioned on both the
previously predicted labels, as well as the keyword of interest.

Unlike the RNN-T model, which can be trained given pairs of input and output
sequences $(\x, \y)$, in order to train the RNN-T model with keyword biasing, we
need to also associate a keyword phrase, $\k$, with the training instance.
We create examples where the keyword, $\k$, is present in $\x$, as well
as examples where the keyword is absent in $\x$ as follows: with probability
$p^\text{kw}$ we uniformly sample one of the words in $\x$ as the keyword, $\k$,
and with probability $1 - p^\text{kw}$ we uniformly sample a word which is not
in $\x$ as the keyword, $\k$.
If we select one of the words in $\x$ as the target, we modify the target labels
$\y$ by inserting a special symbol \texttt{$\eokw$} after the occurence of the keyword.
For example, when training with phoneme targets, for the utterance \texttt{the cat sat},
(which corresponds to the phoneme sequence\footnote{We use X-SAMPA to denote phonemes throughout
the paper.}
\texttt{$[\text{D}~\text{V}~\eow~\text{k}~\text{\{}~\text{t}~\eow~\text{s}~\text{\{}~\text{t}~\eow]$}),
if we sampled \texttt{$\k=$cat} as the keyword, then we would modify the target labels as,
\texttt{$\y=[\text{D}~\text{V}~\eow~\text{k}~\text{\{}~\text{t}~\eow~\eokw~\text{s}~\text{\{}~\text{t}~\eow]$}.
Note that the \texttt{$\eow$} token marks the end of each word token (see Section~\ref{sec:keyword_filler}).
The intuition behind adding the \texttt{$\eokw$} at the end of the keyword
phrase in the transcript, is that it might serve as a marker that the model
should attend to the targets in the keyword phrase.
As a final note, the training and inference algorithms for this model are
similar to the standard RNN-T model.

\section{Baseline Systems}
\label{sec:baselines}
We present two baseline approaches for the task of streaming KWS.
First, we adapt an embedded LVCSR system, designed for efficient real-time
recognition on a wide variety of smartphones, developed in our previous
work~\cite{McGrawPrabhavalkarAlvarezEtAl16}.
Second, we explore ``keyword-filler" models~\cite{Szöke2005} using the acoustic model
component of the LVCSR system.
These approaches are described in the following sections.

\subsection{LVCSR with CTC}
\label{sec:lvcsr}
Our first approach directly uses an embedded LVCSR system developed in our
previous work~\cite{McGrawPrabhavalkarAlvarezEtAl16} to recognize input
utterances; this is followed by a simple confidence estimation scheme in order
to detect a particular keyword of interest.
In particular, we recognize the input utterance, $\x$, and create an n-best list
of hypotheses, denoted as $\mathcal{W}$.
Note that the output vocabulary of the system is limited to 64K words, which
results in a significant number of out-of-vocabulary words during the search
process.
In previous works, e.g., ~\cite{RosePaul90}, the KWS confidence metric is
defined as a likelihood ratio of the keyword model to a background model.
Similar to the approaches, we define a simple confidence metric based on the
n-best list, as follows.
Given an utterance $\x$, we identify the highest probability hypothesis in
$\mathcal{W}$ containing $\k$: $P(\mathbf{w}^+ | \x)$, and the highest
probability hypothesis in $\mathcal{W}$ which does not contain $\k$:
($P(\mathbf{w}^- | \x)$), setting these to 0 if no such hypothesis exists in the
n-best list. We can then compute a confidence metric $C(\x) \in [0, 1]$ as:
\begin{equation}
	C(\mathbf{x}) =
	\frac{P(\mathbf{w}^{+}|\x)}{P(\mathbf{w}^{+}|\x) + P(\mathbf{w}^{-}|\x)}
	\label{eq:confidence}
\end{equation}
Thus, in the case where all n-best entries contain the keyword, the confidence score
is set to one; when none of the entries contain the keyword, the score is set to
zero. This same confidence metric is used for all systems, including the RNN-T
systems presented in this paper.

\subsection{Keyword-Filler Models with CTC}
\label{sec:keyword_filler}
An alternative approach to KWS is through the use of ``keyword-filler"
models~\cite{Szöke2005}, which corresponds to constucting a decoder graph with
two basic paths: the first is a path through the keyword(s), and the second is a
path through a filler (background) that models all non-keyword speech.
We use this approach to create our next set of keyword spotters.

\begin{figure}
  \centering
  \includegraphics[width=0.8\columnwidth]{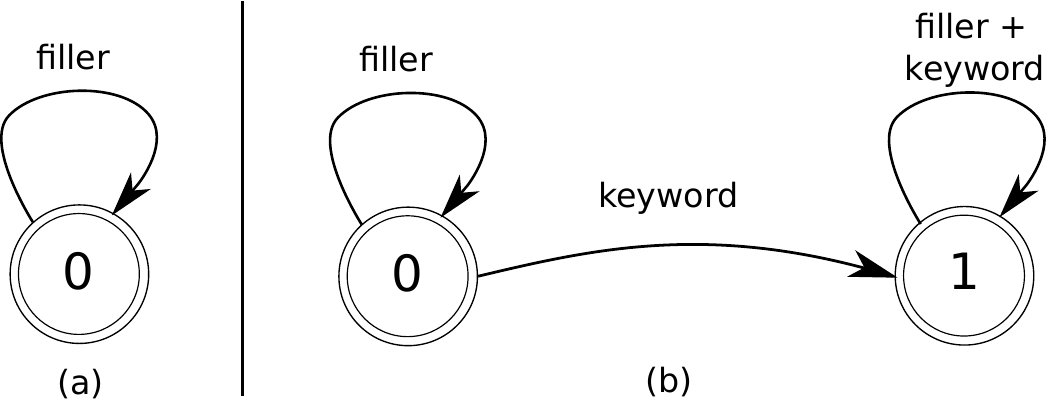}
  \caption{Two decoder graphs representing the building blocks of our baseline
    CTC-based keyword spotters.}
  \vspace{-0.2in}
  \label{two-graphs}
\end{figure}
Instead of defining a single decoder graph with keyword and filler paths, we find it
advantageous to use two decoders on separate graphs as depicted in
Figure~\ref{two-graphs}.
This effectively corresponds to using two beams during decoding: one for the
filler model (Figure~\ref{two-graphs} (a)), and one for the keyword paths,
(Figure~\ref{two-graphs} (b)).
The scores of the most likely paths from each these graph can be used to
estimate $P(\mathbf{w}^- | \x)$ and  $P(\mathbf{w}^+ | \x)$, respectively, which
can be used to generate a confidence score using Equation~\ref{eq:confidence}.

The simplest example of a filler model is a phone loop.
However, we remove all paths from the filler model which contain the keyword's
phones, so that any path containing the keyword must pass through the keyword
model.

In previous work it has been shown that constraining filler models yields
accuracy improvements~\cite{WangS13,chen2015keyword,weintraub1993keyword}.
We therefore explore two variants along these lines.
In the first, we replace the simple phone loops with unweighted word loops
(using the 64k word vocabulary from~\cite{McGrawPrabhavalkarAlvarezEtAl16}),
thus adding in word-level constraints.
In the second, we apply an n-gram phone LM, trained on automatically generated
phonetic transcriptions of the same utterances that are used to train the
word-level LM in~\cite{McGrawPrabhavalkarAlvarezEtAl16}; the number of
parameters in the phone LM is trained to match the number of parameters of the
word LM in~\cite{McGrawPrabhavalkarAlvarezEtAl16}.
In this case, we compose the LM with both the filler and keyword graphs.

In preliminary experiments, we found that a source of false-positives during KWS
with phoneme based models was when a part of word's phonetic transcription
matched that of the keyword.
For example, the keyword \texttt{\small Erica (E~r\textbackslash~@~k~@)} is incorrectly
detected in utterances containing the word,
\texttt{America (@~m~E~r\textbackslash~@~k~@)};
\texttt{\small Marilyn (m~E~r\textbackslash~@~l~@~n)} is incorrectly
detected in utterances containing the word,
\texttt{\small Maryland (m~E~r\textbackslash~@~l~@~n~d)}.
We therefore expanded the phoneme LM by inserting a special symbol \texttt{$\eow$} at
the end of each word's pronunciation when creating training data, e.g.,
\texttt{the cat sat $\to$ D~V~$\eow$~k~\{~t~$\eow$~s~\{~t~$\eow$}.
The \texttt{$\eow$} token is the analog of the space symbol which delimits
words in their graphemic representation; from the long context along with the
\texttt{$\eow$} symbol, the phone LM is expected to implicitly model word-level
dependencies and learn the correct segmentation of a phone sequence into words.
During search, we only consider keywords in between two end-of-word markers, or
between a start-of-sentence marker and an end-of-word marker, in the hypotheses.
For instance, \texttt{\small Erica} would not be false triggered in the
phrase: \texttt{\small In America (I n $\eow$ @ m E r\ @ k @ $\eow$)},
but will correctly trigger when the utterance contains
\texttt{Call Erica (k O l $\eow$ E r\textbackslash~@ k @ $\eow$)}.

The idea of using an end-of-word symbol has also been explored
in~\cite{ZhuangEtAl16}, however the authors added it to the transcript for
training the CTC acoustic model instead. We believe it would be more explicit
and effective to use the symbol for LM training, in which the label
dependencies are modeled directly, whereas in CTC the output targets are
conditionally independent to each other. As is shown in the results below, we
also use the end-of-word symbol for training RNN-T models and find it useful,
where AM and LM are jointly trained.

\section{Experimental Details}
\label{sec:experiments}

\subsection{Data and Evaluation Metric}

Our models are trained on a set of $\sim$22M hand-transcribed anonymized
utterances extracted from Google voice-search traffic, which corresponds to
$\sim$18,000 hours of training data.
In order to improve system robustness to noise and reverberation,
multi-condition training (MTR) data are generated: training utterances are
artificially distorted using a room simulator, by adding in noise samples
extracted from YouTube videos and environmental recordings of daily events.
To further improve robustness to variation in signal loudness, we perform
multi-loudness training by scaling the loudness of each training utterance to a
randomly selected level.

We construct separate development and test sets to measure KWS
performance.
As keyword phrases we consider personal names which contain three or more
syllables (e.g., \texttt{Olivia} or \texttt{Erica}).
The development set consists of 328 keywords, each of which is contained in
$\sim$75 positive utterances, collected from multiple speakers, of the form ``keyword, query'', (e.g.,
\texttt{Olivia, how tall is the Eiffel tower?}).
A set of $\sim$37K negative utterances ($\sim$50 hours in total) are shared
across keywords, which are collected as queries without a keyword, to form the
full development set.
Each keyword is evaluated separately on a set consisting of its own positive
utterances and the shared negative utterances.
A test set is created similarly, with 228 keywords each contained in $\sim$500
positive utterances, and a set of $\sim$20k negative utterances ($\sim$60
hours in total) shared across keywords, which consist of hand-transcribed
anonymized utterances extracted from Google traffic from the domains of
open-ended dictation and voice-search queries.

We evaluate performance in terms of the receiver operating characteristic (ROC)
curve~\cite{MohriCortes04}, which is constructed by sweeping a threshold over
all possible confidence values and plotting false reject (FR) rates against
false alarm (FA) rates. Our goal is to achieve low FR rates while maintaining
extremely low FA rates (e.g. no more than 0.1 false alarms per hour of audio).

Following~\cite{ZhangEtAl12}, we employ a score normalization approach to map
system confidence score at the utterance level for a keyword to the probability
of false alarm (pFA) for that keyword, which allows us to use a single
consistent score for all keywords and set the decision threshold reliably.
A confidence-score-to-pFA mapping is estimated from the development set, and
applied to both the development and the test sets.
All ROC curve results in this work are plotted after the score normalization.

\subsection{Model Details}

The input acoustic signal is represented with 80-dimensional log-mel filterbank
energies, computed with a 25ms window, and a 10ms frame-shift.
Following previous work~\cite{SakSeniorRaoEtAl15}, we stack three consecutive
frames and present only every third stacked frame as input to the encoder.
The same acoustic frontend is used for all experiments described in this work.

The CTC acoustic model (AM) consists of 5 layers of 500 LSTM cells, that predict
context-independent phonemes as output targets.
The system is heavily compressed, both by
quantization~\cite{AlvarezPrabhavalkarBakhtin16}, and by the application of
low-rank projection layers with 200 units between consecutive LSTM
layers~\cite{PrabhavalkarAlsharifBruguierMcGraw17}.
The AM consists of 4.6 million parameters in total.
The model is first trained to optimize the CTC objective
function~\cite{GravesFernandezGomezEtAl06} until convergence.
Once CTC-training is complete, the model is discriminatively sequence-trained to
optimize expected word errors by minimizing word-level, edit-based, minimum Bayes
risk (EMBR) proposed recently by Shannon~\cite{Shannon17}.

The encoder networks used in all RNN transducer models are identical in size and
configuration to the encoder used in the CTC model (without the softmax output layer).
During the training of an RNN transducer, the weights from the encoders are
initialized from a pre-trained CTC model, since this was found to significantly
speed up convergence, following which the weights are trained jointly with the
rest of the network.
For the RNN-T model that is trained to directly output grapheme targets, the CTC
model used for initialization is also trained to predict graphemes.
The grapheme inventory includes the 26 lower-case letters
(\texttt{a-z}), the numerals (\texttt{$0$-$9$}), a label representing `space'
(\texttt{$\spc$}), and punctuation
symbols (e.g., the apostrophe symbol (\texttt{'}), hyphen (\texttt{-}), etc.).

The prediction network used in the RNN transducer models, both with and without
attention, consists of a single layer of 500 LSTM cells with coupled input and
forget gate (CIFG)~\cite{GreffEtAl16}, and the joint network
consists of a single feed-forward layer of 500 units with a $\tanh$ activation
function, as described in Section~\ref{sec:rnnt}.
The decoder network (including prediction network and the joint network) has 1.5
million parameters in total.

The RNN transducer models are decoded using a beam-search
algorithm~\cite{Graves12}, where at most 50 highest scoring candidates are
retained at every step during decoding.
In general, the output posterior distribution of sequence-to-sequence models
like RNN-T is peaky (i.e., low entropy); such over-confidence is typically
suboptimal for keyword spotting, since diversity in hypotheses is critical to
reduce the number of false rejects.
We find that smoothing the output posteriors with a temperature $\tau$,
i.e. mapping each posterior to its $\tau$-th root and renormalizing them,
can help improve KWS performance significantly.
The optimal temperature value is determined by tuning on the development set; we
set $\tau=2.0$ for all RNN-T models without attention, and $\tau=2.2$ for the
ones with attention. However smoothing the output posteriors of the CTC acoustic
model does not help, possibly because it does not combine well with the LM.

The language model (LM) used in the keyword-filler model with CTC is trained to
predict phoneme targets on the same $\sim$22M utterances used for training RNN-T
models.
The LM is pruned to $\sim$1.5-million 6-grams using entropy pruning, similar to
the number of parameters in the decoder of our RNN-T models.
We choose $n=6$ which is optimized from the development set.

The LM for our embedded LVCSR system is a standard word-level 5-gram, which
is trained on a larger corpus with $\sim$100M automatically-transcribed
anonymized utterances extracted from Google voice-search traffic.
This LM is also pruned to $\sim$1.5-million n-grams using entropy pruning.
The vocabulary is limited to 64K words, allowing us to shrink the data
structures used to maintain the LM~\cite{McGrawPrabhavalkarAlvarezEtAl16}.
Note that the fixed vocabulary results in out-of-vocabulary keywords on the
development and test sets.
Utterances are decoded with a heavily pruned version of the LM in the
first-pass, while rescoring with the full LM on-the-fly, thus allowing us to
reduce the size of the decoder graph used in the first-pass.

All models are trained using asynchronous stochastic gradient
descent~\cite{DeanCorradoMongaEtAl12}, and are implemented in
TensorFlow~\cite{AbadiAgarwalBarhamEtAl15}.

\section{Results}
\label{sec:results}

\subsection{Baselines}
\begin{figure}
	\centering
	\includegraphics[width=0.4\textwidth]{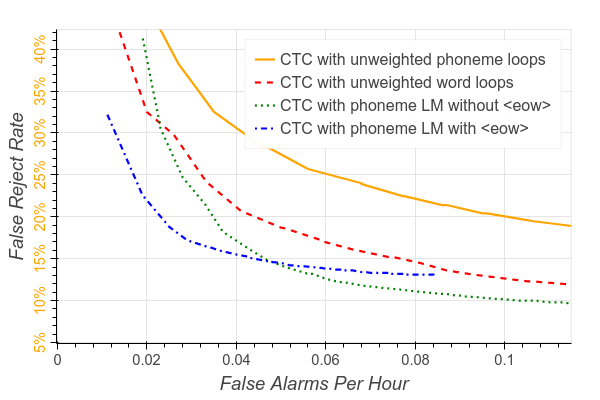}
	\caption{Comparison among multiple CTC baseline systems on the
	test set.}
	\label{fig:ctc}
	\vspace{-0.2in}
\end{figure}
The performance of our CTC-trained ``keyword-filler'' baseline models is shown
in Figure~\ref{fig:ctc}.
As can be seen, we find that a phoneme language model is important for
a keyword-filler system even with a strong CTC model for extremely low FA rates
($\leqslant$ 0.05 FAs per hour).
The effect of the language model can be seen by comparing
different levels of constraints added in the keyword-filler graphs.

CTC with unweighted phoneme loops allows for arbitrary phoneme paths in the
graph to be treated as \emph{equally likely}, thus entirely relying on the CTC
model to recognize the keyword from the background, which performs the worst.
Adding word constraints in the graph, albeit without weights, helps to improve
performance since it eliminates many confusable paths that correspond to invalid
words.
Note that in this case, we can add the keyword phrase into the vocabulary for
the search since the word loop filler models are unweighted.

The addition of a phoneme language model without the \texttt{$\eow$} token helps
to recognize phoneme sequences in context, but does not account for word
constraints.
As described in Section~\ref{sec:keyword_filler}, this model has an increased
number of false triggers (e.g., the keyword \texttt{Erica} is detected
incorrectly in utterances containing the word \texttt{America}).
The addition of \texttt{$\eow$} to the phoneme language models, however,
significantly improves performance over the other baseline systems.

For reference, a KWS system constructed from the embedded LVCSR system, as
described in Section~\ref{sec:lvcsr}, achieves an FR rate of 29.8\% at 0.05 FAs
per hour on the test set for only in-vocabulary keywords (196 out of total 228
keywords), while the best CTC system above achieves 13.4\% on the same set.

\subsection{RNN-T Models with Graphemes and Phonemes Targets}
\begin{figure}
	\centering
	\includegraphics[width=0.4\textwidth]{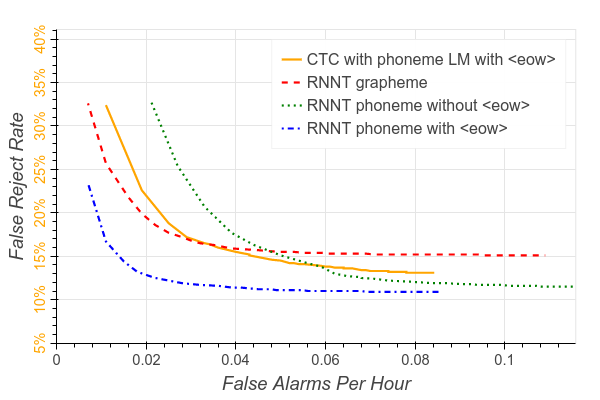}
	\caption{A comparison of the various RNN-T systems against the best
	performance CTC baseline on the test set.}
	\label{fig:rnnt}
	\vspace{-0.2in}
\end{figure}
\begin{figure*}
	\centering
	\begin{subfigure}[b]{0.48\textwidth}
		\includegraphics[width=\textwidth]{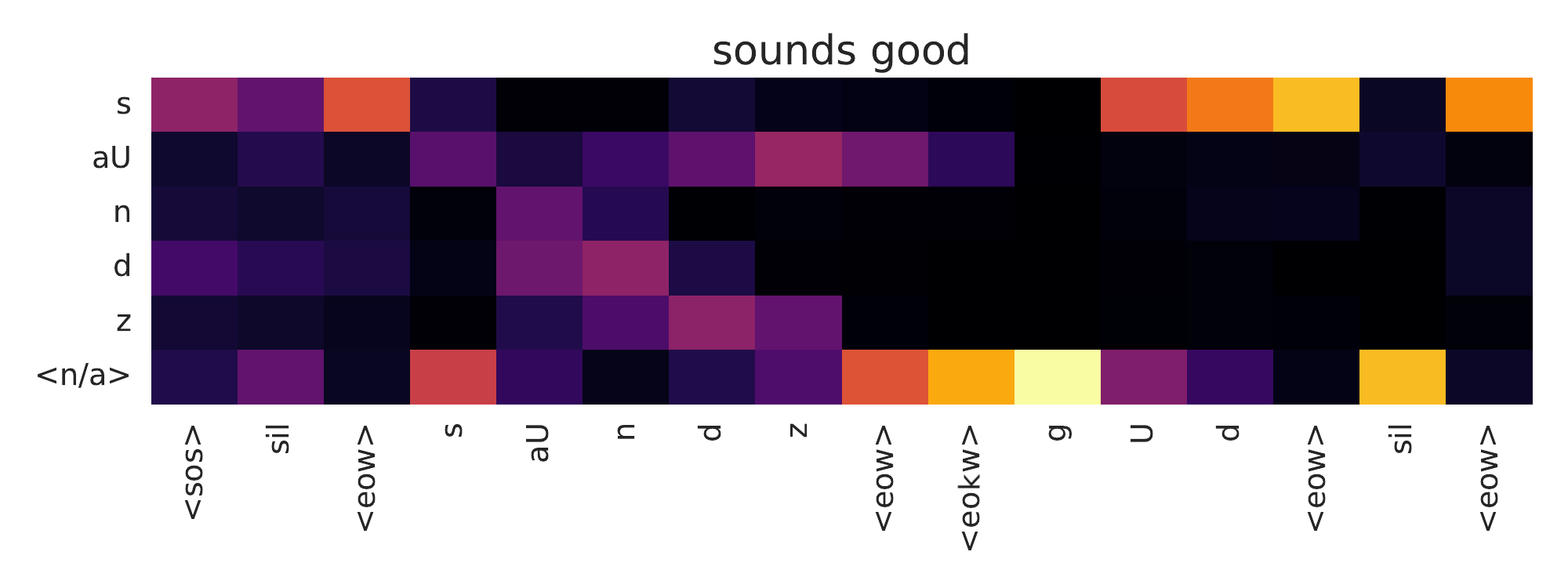}
		\caption{Attention matrix of a positive utterance for the
		keyword ``sounds'', with the transcript ``sounds good''.}
		\label{fig:attn_pos}
	\end{subfigure}
	~
	\begin{subfigure}[b]{0.48\textwidth}
		\includegraphics[width=\textwidth]{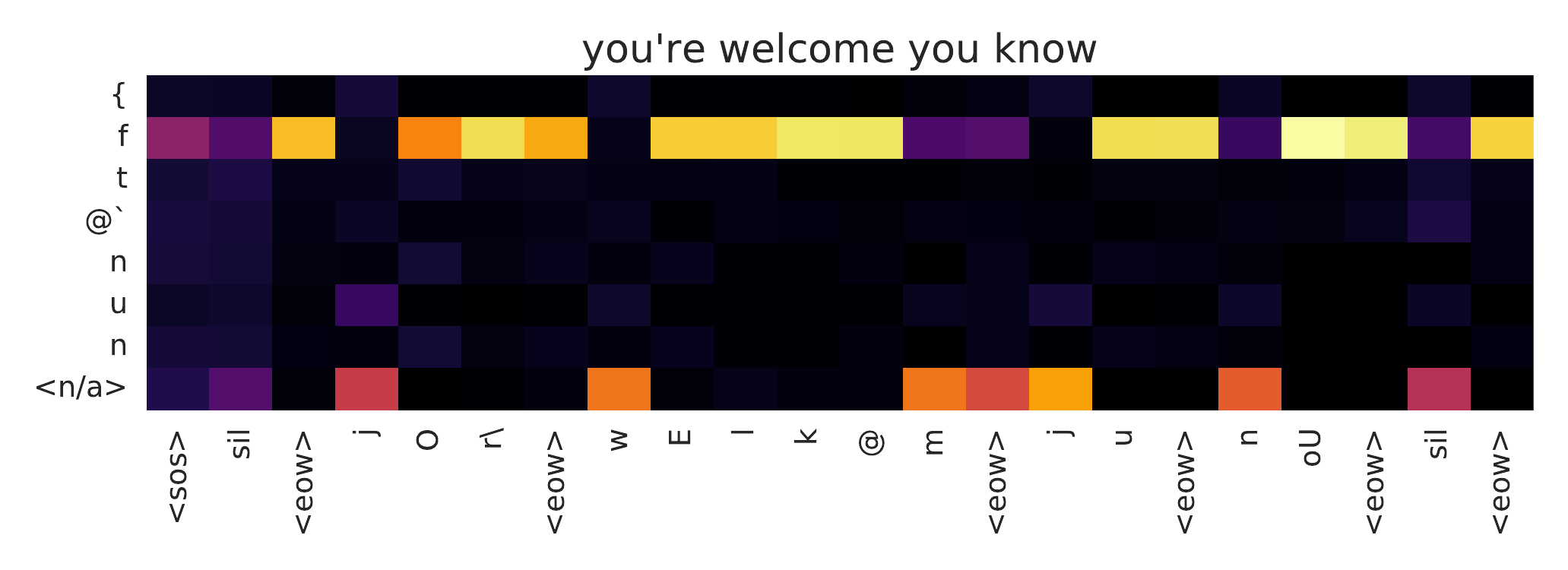}
		\caption{Attention matrix of a negative utterance for the keyword ``afternoon'',
		with the transcript ``you're welcome you know''.}
		\label{fig:attn_neg}
	\end{subfigure}
	\caption{Attention matrices for two representative utterances computed
	by the RNN-T phoneme system with keyword biasing. The Y-axis corresponds
	to targets $k_1, \cdots, k_{M+1}$ in the keyword $\k$. The X-axis
	corresponds to the expected sequence of phoeneme targets given the
	utterance transcript. The entry at row $j$ and column $u$ corresponds to $\alpha_{j, u}$ in
	Equation~\ref{eq:att_weights}, with values in each column summing up to
	1. Brighter colors correspond to values closer to 1, while darker colors
	correspond to values closer to 0.}
	\label{fig:attn_plot}
	\vspace{-0.2in}
\end{figure*}
Compared to CTC with a phoneme n-gram LM, an RNN-T model with phoneme targets
jointly trains an acoustic model component and a language model component in a
single all-neural system.
As can be seen from Figure~\ref{fig:rnnt}, an RNN-T phoneme
model (with \texttt{$\eow$}) outperforms the best CTC baseline.
If the \texttt{$\eow$} token is not used, however, the RNN-T phoneme system has
significantly higher false alarms as explained in
Section~\ref{sec:keyword_filler}.

The RNN-T system trained to predict grapheme targets performs worse than
the one trained with phoneme targets.
We conduct an analysis to determine the cause of this performance degradation
and found that it is partly due to variant orthographic representations of some
of the keyword phrases: e.g., the keyword \texttt{kathryn} is encountered very
rarely in the training data, and as a result the RNN-T model typically
recognizes these examples as \texttt{catherine}, which is more common in the
training data.
We therefore considered a variant system where we replace each keyword with the
most frequent orthographic representation (as determined by its unigram
probability) during the search.
This technique significantly improves false reject rates for the RNN-T grapheme
system from 15.5\% to 14.0\% at 0.05 FAs per hour; however this system was still
worse than the RNN-T phoneme model, which achieves an FR rate of 11.1\% at 0.05
FAs per hour.

\subsection{RNN-T with Keyword Biasing}
\begin{figure}
	\centering
	\includegraphics[width=0.4\textwidth]{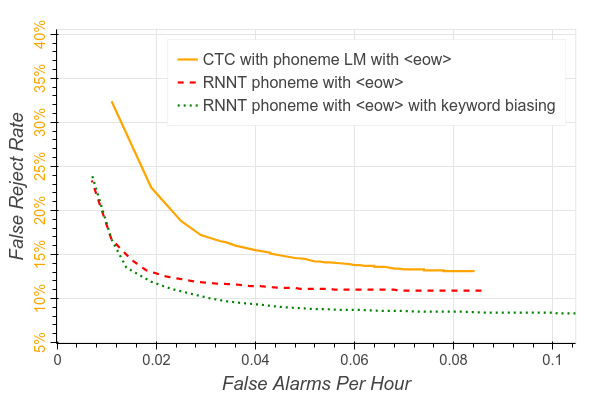}
	\caption{A comparison of the RNN-T phoneme model with keyword biasing
	against the best CTC baseline and the RNN-T phoneme system without
	biasing on the test set. All systems use the \texttt{$\eow$} token.}
	\label{fig:biasing}
	\vspace{-0.1in}
\end{figure}
\begin{figure}
	\centering
	\includegraphics[width=0.4\textwidth]{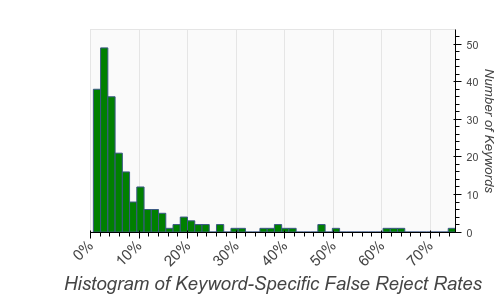}
	\caption{Histogram of keyword-specific false reject rates for the RNN-T
	phoneme system with keyword biasing at 0.05 FAs per hour, plotted for
	the keywords on the test set.}
	\label{fig:fr_hist}
	\vspace{-0.2in}
\end{figure}
We train an RNN-T phoneme system with \texttt{$\eow$} and \texttt{$\eokw$} labels,
by setting $p^\text{KW}=0.5$, determined by tuning on the development set.
As is shown in Figure~\ref{fig:biasing}, adding attention-based keyword biasing
to an RNN-T phoneme system improves the overall performance significantly.
The final results are reported on the test set, where CTC, RNN-T phoneme and
RNN-T phoneme with biasing achieve 14.5\%, 11.1\% and 8.9\% false reject rates
respectively at 0.05 FAs per hour.

We also plot a histogram of the FR rates across keywords at a threshold
corresponding to 0.05 FAs per hour for the RNN-T phoneme system with
keyword biasing in Figure~\ref{fig:fr_hist}.
As can be seen in the figure, most of the keywords have low FR rates
in the 0--15\% range, with only a few outliers.

Finally, in Figure~\ref{fig:attn_plot} we plot representative examples of the
attention weights $\alpha_{j, u}$ computed by the attention model during
inference on a positive (Figure~\ref{fig:attn_plot} (a)) and a negative
(Figure~\ref{fig:attn_plot} (b)) utterance extracted from the training data.
These plots were generated by feeding as input the expected target label
sequence (i.e., the labels are not determined by a beam-search decoding).

As can be seen in the figure, when decoding the positive utterance, the
attention weights are concentrated on the first target.
When the model begins to predict the phonemes corresponding to the keyword
(\texttt{sounds (s aU n d z)}), the attention weights are focussed on
consecutive keyword targets, as revealed by the prominent diagonal pattern
(although admittedly, the model also appears to attend to other keyword targets
during this process).
We also note the prominent attention weight assigned to the \texttt{$\na$} label
after the keyword has been detected.

In the case of the negative utterances, however, the attention does not evolve
diagonally across the labels, but is instead spread across the second keyword
target (i.e., the initial part of the hotword), and the \texttt{$\na$} label.

\section{Conclusions}
\label{sec:conclusions}
In this work, we developed streaming keyword spotting systems using
a recurrent neural network transducer, a sequence-to-sequence model that
jointly trains acoustic and language model components. We proposed
a novel techinque which biases the RNN-T system towards a specific keyword
of interest based on an attention mechanism over the keyword. In
experimental evaluations, we find that our RNN-T system trained with phoneme targets
performs significantly better on keyword spotting than a strong CTC-based
keyword-filler baseline which is augmented with a phoneme n-gram LM.
We also find that the proposed biasing techique provides further gains over the
vanilla RNN-T model.

\bibliographystyle{IEEEbib}
\bibliography{main}

\end{document}